\pdfoutput=1

\documentclass[11pt]{article}

\usepackage{EMNLP2022}

\usepackage{times}
\usepackage{latexsym}

\usepackage[T1]{fontenc}

\usepackage[utf8]{inputenc}

\usepackage{microtype}

\usepackage{inconsolata}

\usepackage{booktabs}
\usepackage{graphicx}
\usepackage{multirow}
\usepackage{amssymb}
\usepackage{mathtools}
\usepackage[export]{adjustbox}
\usepackage{subcaption}

\title{Visual Named Entity Linking: A New Dataset and A Baseline}

\author{Wenxiang Sun,\ \ Yixing Fan,\ \ Jiafeng Guo\thanks{{ }{ }Corresponding authors.},\ \  Ruqing Zhang,\ \ Xueqi Cheng\\
  CAS Key Lab of Network Data Science and Technology, Institute of Computing Technology, \\
  Chinese Academy of Sciences, Beijing, China\\
  University of Chinese Academy of Sciences, Beijing, China\\
  \texttt{\{sunwenxiang20s,fanyixing,guojiafeng,zhangruqing,cxq\}@ict.ac.cn}}

\begin{document}
\maketitle
\begin{abstract}
Visual Entity Linking (VEL) is a task to link regions of images with their corresponding entities in Knowledge Bases (KBs), which is beneficial for many computer vision tasks such as image retrieval, image caption, and visual question answering. While existing tasks in VEL either rely on textual data to complement a multi-modal linking or only link objects with general entities, which fails to perform named entity linking on large amounts of image data. In this paper, we consider a purely \textbf{V}isual-based \textbf{N}amed \textbf{E}ntity \textbf{L}inking (VNEL) task, where the input only consists of an image. The task is to identify objects of interest (i.e., visual entity mentions) in images and link them to corresponding named entities in KBs. Since each entity often contains rich visual and textual information in KBs, we thus propose three different sub-tasks, i.e., visual to visual entity linking (V2VEL), visual to textual entity linking (V2TEL), and visual to visual-textual entity linking (V2VTEL). In addition, we present a high-quality human-annotated visual person linking dataset, named WIKIPerson. Based on WIKIPerson, we establish a series of baseline algorithms for the solution of each sub-task, and conduct experiments to verify the quality of the proposed datasets and the effectiveness of baseline methods. We envision this work to be helpful for soliciting more works regarding VNEL in the future. The codes and datasets are publicly available at \url{https://github.com/ict-bigdatalab/VNEL}.
\end{abstract}

\begin{figure}[t!]
  \centering
  \setlength{\abovecaptionskip}{0.cm}
  \includegraphics[width=\linewidth]{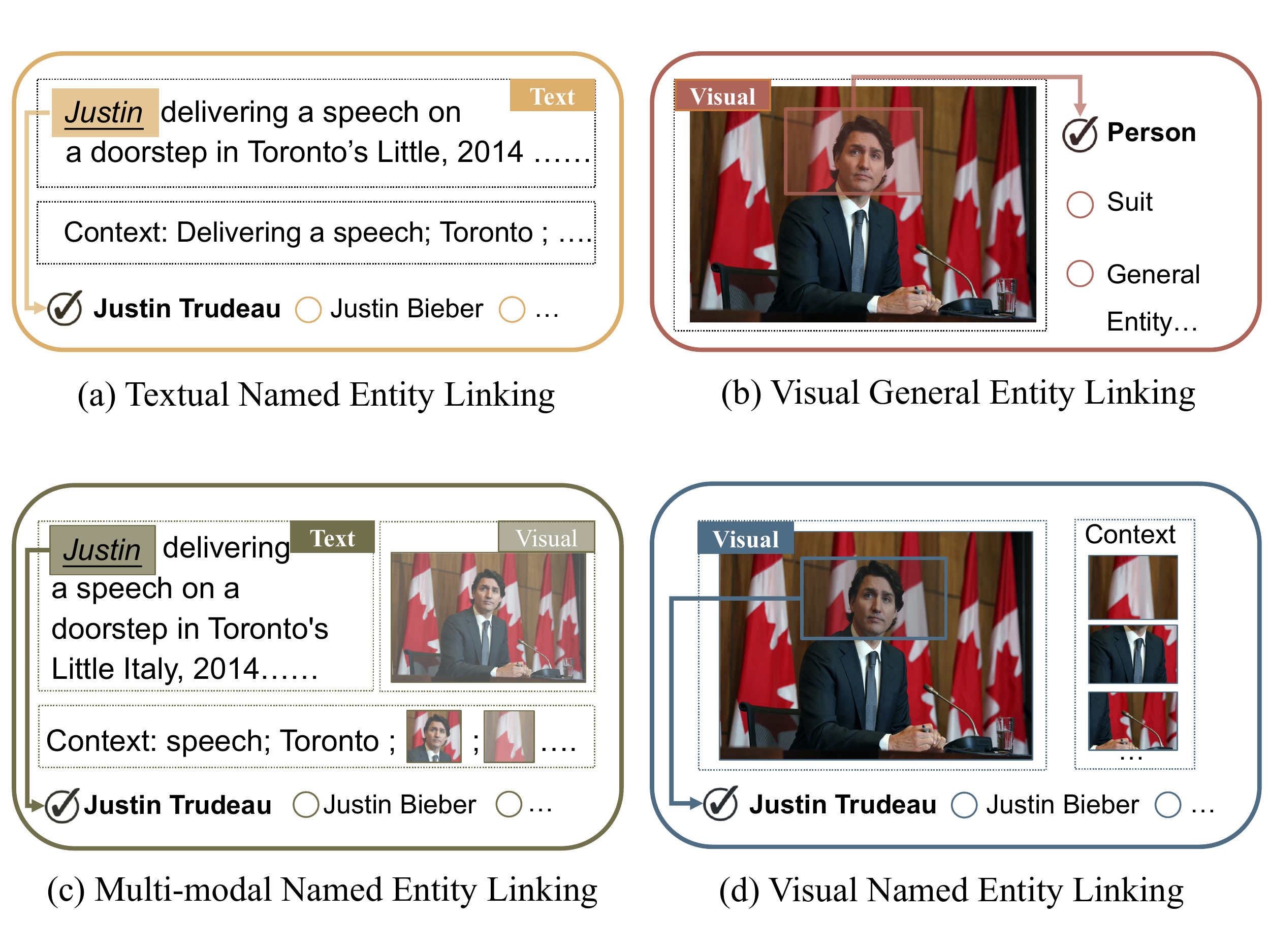}
  \caption{Different categories of Entity Linking. VNEL is a task to identify images individually without any text input and link visual mentions to specific named entities in KBs.}
  \label{IEL}
  \vspace{-0.5cm}
\end{figure}

\begin{table*}[t!] \scriptsize
\setlength\tabcolsep{3.2pt}%
  \begin{tabular}{l|ccclcccc}
    \toprule
    \midrule
    Dataset & Multi-modal & Entity-aware & Entity-labeled & Modality & KB & Source & Lang
 & Size\\
    \midrule
    AIDA\cite{AIDA} & &  \checkmark &  \checkmark & $T^m \rightarrow T^e$ & Wikipedia& News & en & 1K docs\\
    Flicker30K\cite{Fliker} & \checkmark &  &  &  & & social media  & en & 30k images\\
    BreakingNews\cite{BreakingNews} &\checkmark & \checkmark & & & & News  & en & 100k images\\
    SnapCaptionsKB\cite{snap} &\checkmark & \checkmark & \checkmark & $T^m + V \rightarrow T^e $ & Freebase & Social Media & en & 12K captions \\
    WIKIDiverse\cite{WikiDiverse} &\checkmark & \checkmark & \checkmark & $T^m + V \rightarrow T^e, V^e$ & Wikipedia & News & en & 8K captions\\
    \midrule
    \multirow{3}*{WIKIPerson}&  &  &  & $V^m \rightarrow V^e$  &   &   &   &  \\
 ~ & \checkmark &  \checkmark &  \checkmark & $V^m \rightarrow T^e $  & Wikipedia & News & en & 50k Images\\
 ~ &  &  &  & $V^m \rightarrow V^e, T^e$  &   &   &   &  \\
  \bottomrule
\end{tabular}
\caption{The public related dataset of WIKIPerson. $T^m$, $T^e$, $V^m$, $V^e$, and $V$ represent textual mention, textual entity, visual mention, visual entity, and visual information, respectively.}
\label{tab:dataset}
\end{table*}

\section{Introduction}
An in-depth understanding of visual content in an image is fundamental for many computer vision tasks. 
VEL \citep{VEL,maigrot2016mediaeval} is a task to put the image understanding to the entity-level. For example, given an image of the debate between Trump and Hillary, the goal of VEL is not only to recognize the region of Trump and Hillary, but also to link them to the correct entity in KBs (e.g., Wikidata \citep{WIKIdata}, DBpedia \citep{DBpedia}, or YAGO \citep{YAGO}). 
Just as the significance of textual entity linking for many NLP tasks such as Information Extraction and Information Retrieval \citep{EL_survey}, visual tasks, such as image retrieval \citep{imageRetrieval} and image caption \citep{IC_1}, would also benefit from entity-level fine-grained comprehension of images.

In recent years, VEL has been given increasing attention. Early works~\citep{VEL,VEL_Pre} try to link objects in images with general entities, e.g., `Person' and `Suit', in KBs as is described in Figure~\ref{IEL}(b). Apparently, these works are restricted to the coarse-level entity linking and fail to distinguish objects within the same class. Besides, there are also some works that make use of deep image understanding to link objects with named entities in KBs~\citep{VEL_Linker,VEL_MML,VEL_General,gan2021multimodal}. However, they generally require detailed entity mention information in text, which plays a vital role via multi-modal entity linking as shown in Figure~\ref{IEL}(c). We argue that all the above tasks fail to process the named entity linking well for images without any text annotations, which is often the case in social media platforms.

In this work, we consider a purely \textbf{V}isual-based \textbf{N}amed \textbf{E}ntity \textbf{L}inking (VNEL) task, which is described in Figure~\ref{IEL}(d). Given an image without textual description, the goal is to link the visual mention in the image with the whole image as the context to the corresponding named entity in KBs. Considering the format of entity in KBs, such as textual descriptions, images, and other structured attributes, we further introduce three sub-tasks according to the type of entity context, i.e., the visual to visual entity linking (V2VEL), visual to textual entity linking (V2TEL), and visual to visual-textual entity linking (V2VTEL). We believe these tasks could put forward higher requirements and more detailed granularity for image understanding, cross-modal alignment, and multi-modal fusion.

Following the definition of VNEL, currently public available EL datasets may not fit for our research, as they either only focus on textual modality or lack of detailed annotations for entity information in each image. As a result, we release a new dataset called WIKIPerson. The WIKIPerson is a high-quality human-annotated visual person linking dataset based on Wikipedia. Unlike previously commonly-used datasets in EL, the mention in WIKIPerson is only an image containing the PERSON entity with its bounding box. The corresponding label identifies an unique entity in Wikipedia. For each entity in Wikipedia, we provide textual descriptions as well as images to satisfy the need of three sub-tasks.

In the experiments, we benchmark a series of baseline models on WIKIPerson under both zero-shot and fine-tuned settings. In detail, we adopt a universal contrastive learning framework to learn a robust and effective representation for both mentions and entities. Experimental results show that existing models are able to obtain a reasonably good performance on different VNEL tasks, but there is still a large room for further enhancements.

\begin{figure}[t]
  \centering
  \setlength{\abovecaptionskip}{0.5cm}
  \includegraphics[width=\linewidth]{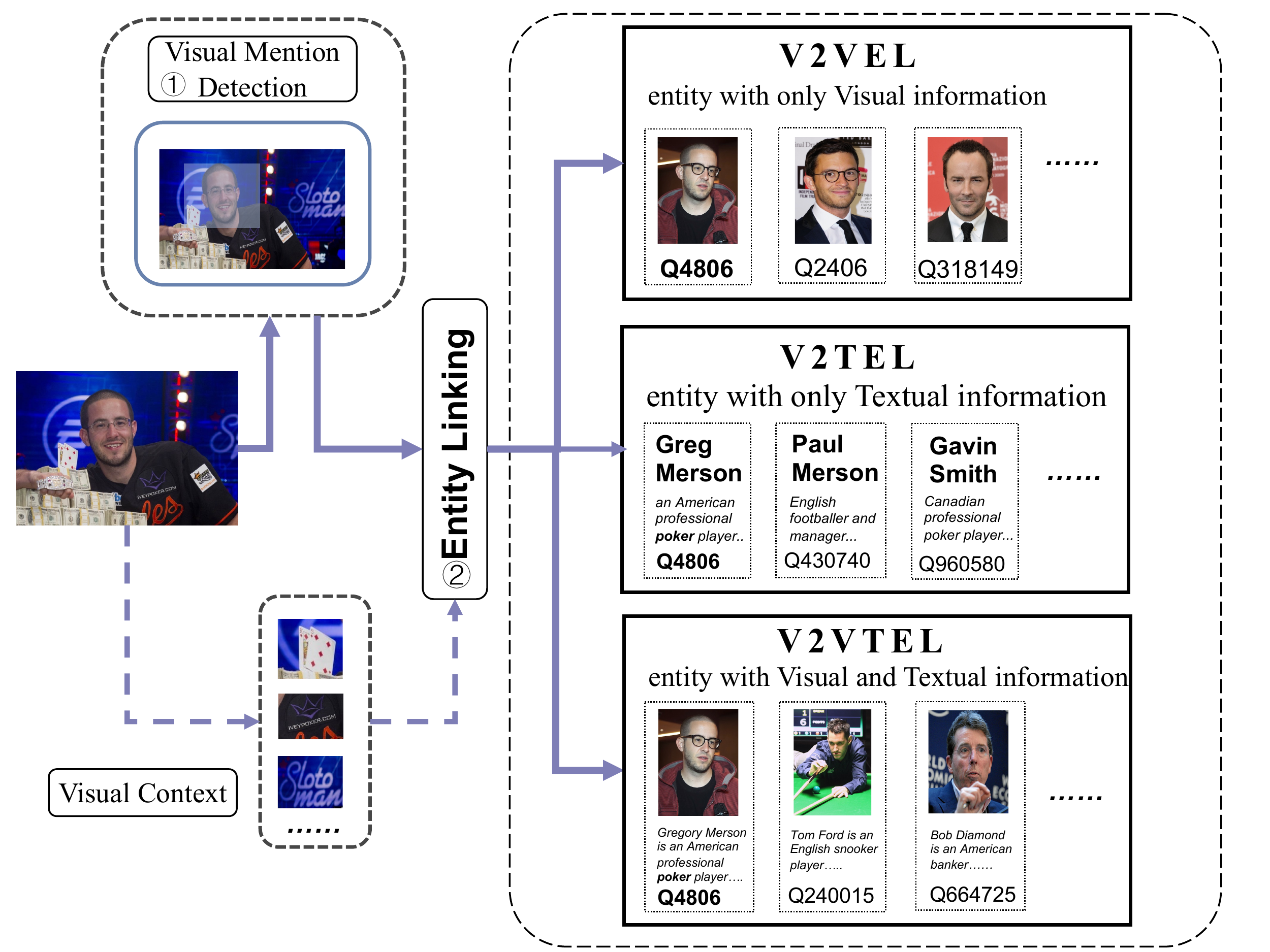}
  \caption{VNEL with its three sub-tasks.}
  \label{three}
  \vspace{-0.5cm}

\end{figure}

\section{The Visual Named Entity Linking Task}
This section first presents a formal definition of the task. Then we introduce the complete building procedure of the human-annotated dataset, which covers a wide variety of Wikipedia person entities for further research. Finally, an in-depth data analysis will be elaborated on in detail.

\subsection{Definition of VNEL and Three Sub-tasks}

VNEL takes an image as input and extracts bounding boxes around objects, and then links them to entities in KBs. More precisely, given an image $I$, all visual mentions $V^m$, which are regions of the image, are firstly recognized with a bounding box. Then, all visual mentions $V^m$ are linked with the corresponding entity $e$ in knowledge base $E$. The visualized process of the VNEL task is shown in Figure \ref{three}, which often consists of two stages, namely the visual mention detection stage and the visual entity linking stage. In this work, we follow existing works \citep{EL_g_1, EL_g_4} to pay attention to the visual entity linking stage.

Generally, each entity $e_i \in E$ is often characterized with rich textual and visual descriptions, and each modality of the description can provide sufficient information for visual entity linking. To make the task more clearly presented, we further decompose the VNEL task into three sub-tasks according to the type of description used for the entity. In the first place, only the visual description $V_{e_i}$ of the entity can be used in the visual entity linking stage, which we denote as the V2VEL sub-task. The core of V2VEL is to match two visual objects. It is worth noting that entities in KB may contain more than one image. To simply this, we take the first image of $e_i$ as $V_{e_i}$, and leave the multiple images per entity as the future work. 
In the second place, only the textual description $T_{e_i}$ of the entity is used in the visual entity linking stage, which we denote as the V2TEL sub-task. The V2TEL task aims to evaluate the ability in image-text matching, central to cross-modal entity linking. Finally, both the visual description and the textual description $(V_{e_i},T_{e_i})$ of the entity could be employed to link the visual mention, which we denote as the V2VTEL sub-task. The V2VTEL task could leverage both textual and visual modality to complement each other in linking visual mentions.

Formally, let $e_i$ represent the $i^{th}$ entity in KB with corresponding visual description $V_{e_i}$ or textual description $T_{e_i}$ and the whole image can be seen as visual context $V^c$. As a result, three sub-tasks of the VNEL can be formulated as the following respectively:

\setlength{\abovedisplayskip}{-0.15cm}
\setlength{\belowdisplayskip}{0.15cm}
$$
\begin{gathered}
\underset{V \rightarrow V}{e^{*}(m)}=\underset{e_{i} \in E}{\arg \max } \Phi^{\alpha}\left(V^{m},  V_{e_i} \mid V^{c}\right), \\
\underset{V \rightarrow T}{e^{*}(m)}=\underset{e_{i} \in E}{\arg \max } \Phi^{\beta}\left(V^{m}, T_{e_i} \mid V^{c}\right), \\
\underset{V \rightarrow V+T}{e^{*}(m)}=\underset{e_{i} \in E}{\arg \max } \Phi^{\gamma}\left(V^{m}, (V_{e_i}, T_{e_i}) \mid V^{c}\right),
\end{gathered}
$$
where $ \Phi$ represents the value of the score function between a mention and an entity.

\subsection{Dataset Setups of WIKIPerson}
\begin{figure}[t!]
  \centering
  \setlength{\abovecaptionskip}{0.3cm}
  \includegraphics[width=\linewidth]{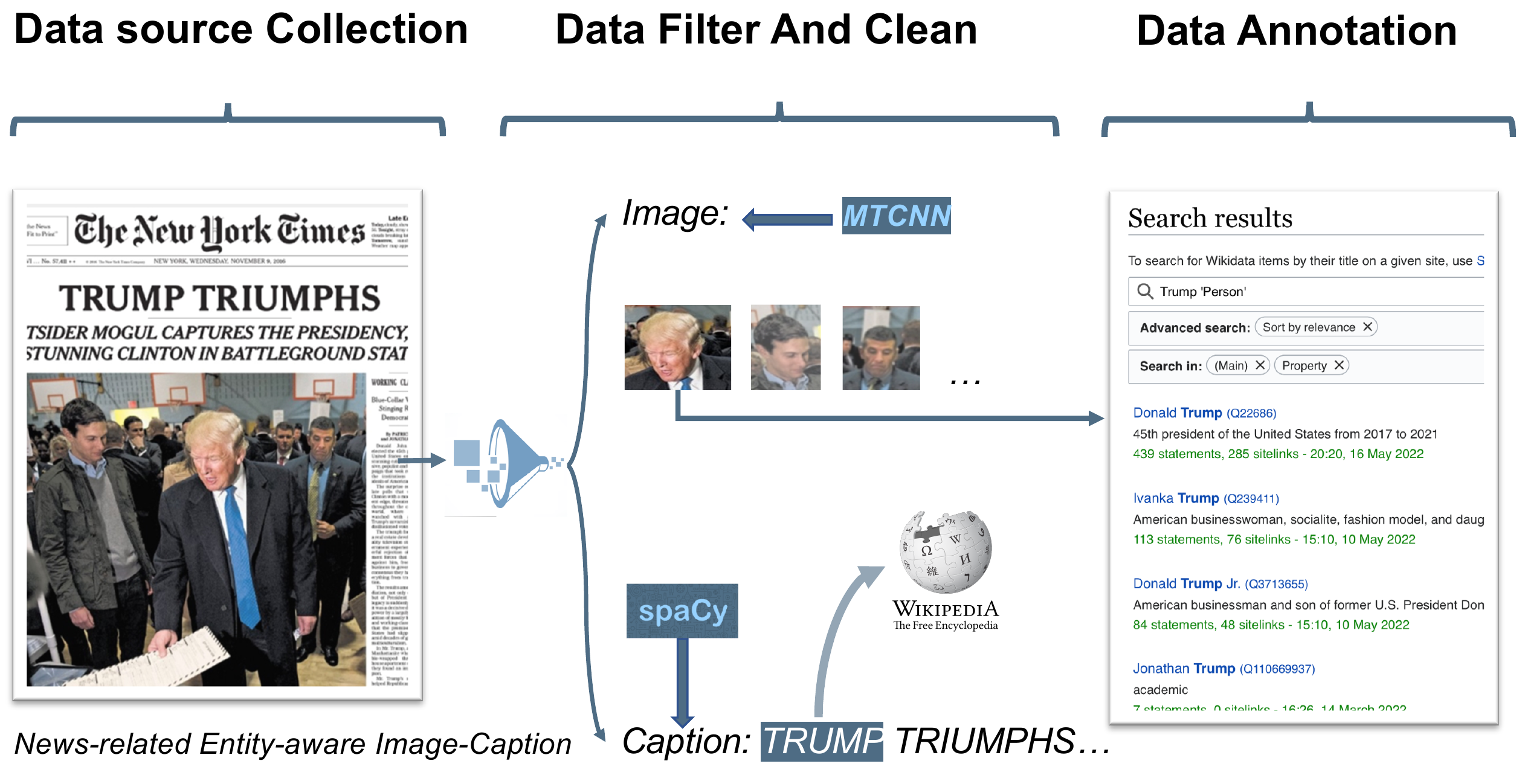}
  \caption{The procedure of building WIKIPerson.}
  \vspace{-0.3cm}
  \label{procecdure} 
\end{figure}

To facilitate research on VNEL, we introduce WIKIPerson, a benchmark dataset designed for linking person in images with named entities in KB. The dataset building process is shown in Figure~\ref{procecdure}, which consists of three main steps. We firstly select the data source to build the input image collection, and then filter and clean the collection to obtain a high-quality dataset. Finally, we annotate each image by several experienced annotators. In the following, we will describe each step in detail.

\subsubsection{Data Source Collection}
For the source of data, we follow existing works~\cite{BreakingNews,NYTimes800k,VisualNews,GoodNews} to use News collections, since the content of images in News collection often contains many named entities at a higher degree of specificity, e.g., specific people, which convey key information regarding the events presented in the images.
In this paper, we choose VisualNews~\footnote{https://github.com/FuxiaoLiu/VisualNews-Repository}, which has the largest data scale with 1.2 million image-text pairs among them as the original data source. In addition, VisualNews covers diverse news topics, consisting of more than one million images accompanied by news articles, image captions, author information, and other metadata.
All these additional metadata could help us in the subsequent entity annotation procedure. However, only images and annotated mentions with bounding boxes are available in all VNEL sub-tasks.

For the knowledge base, we employ the commonly-used Wikipedia as back-end, consisting of a wide range and abundant information of entities. Specifically, we crawl the first image of each entity from wiki commons as the visual description and the text information from Wikipedia as the textual description, respectively.

\subsubsection{Data Filter and Clean}
In this work, we pay our attention to PERSON mentions in images since person is the most common named entity, and leave the research on other entity types for future work.
For this purpose, we keep only images with PERSON mentions from the news collection, and remove non-PERSON entities from the KB. 
Specifically, for each image-caption pair in the news collection, we take Spacy to analyze the text caption and filter out the corresponding data without any PERSON entities. 
Moreover, we leverage the MTCNN model~\cite{MTCNN}, which is the state-of-the-art face detection model, to check the number of PERSON mentions in each image. Then, we select images with the number of person mentions less than 4 to reduce the complexity of the task. Lastly, we remove repeated and blurred images to keep the quality of the dataset. 

\begin{figure}[t!]
  \centering
  \setlength{\abovecaptionskip}{0.3cm}
  \includegraphics[width=1\linewidth]{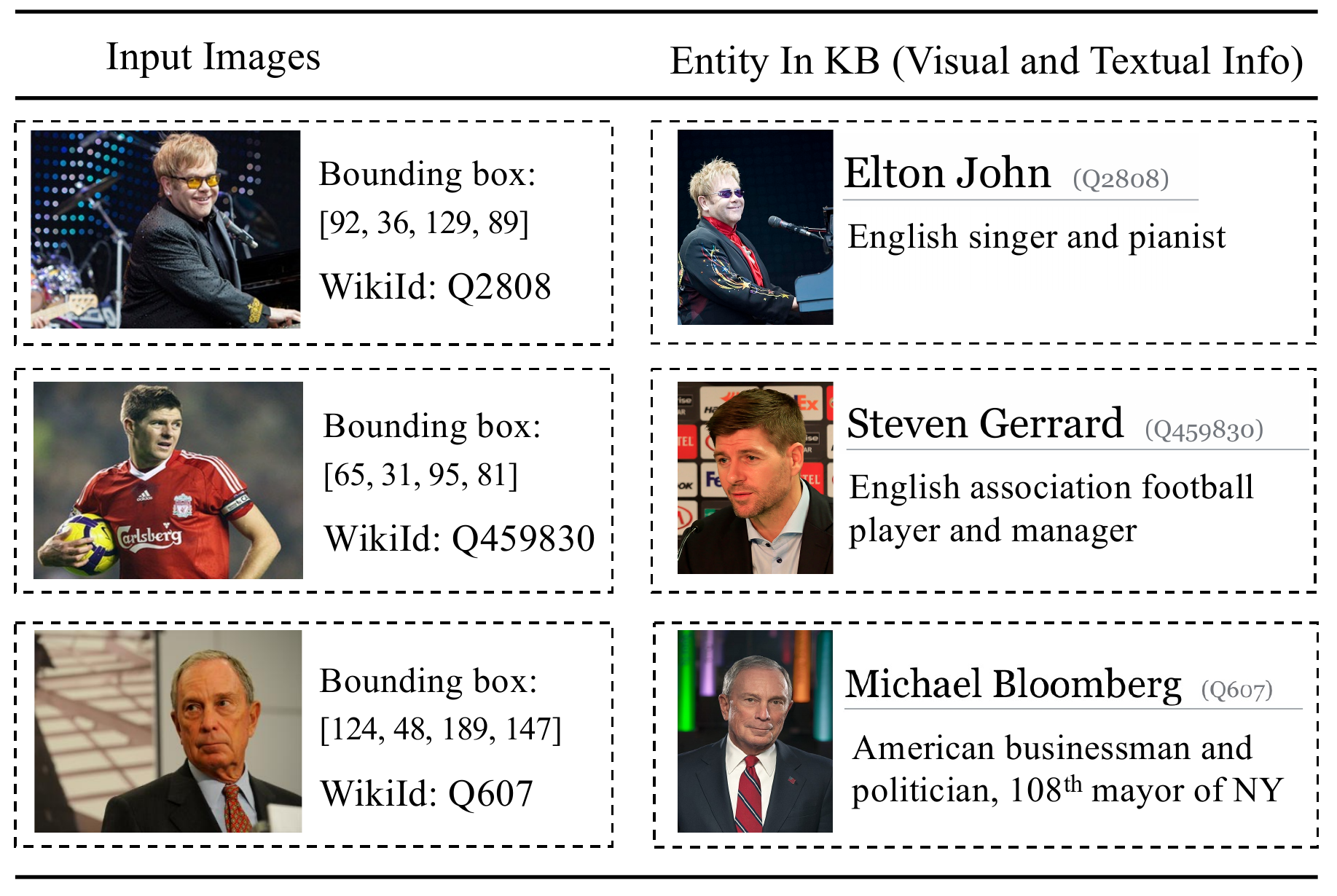}
  \caption{Examples of the WIKIPerson dataset. \textbf{Left}: An image and its mention's bounding box with WikiId, which represents an unique entity in Wikipedia. \textbf{Right}: The ground truth entity in KB with both visual and textual information.}
  \vspace{-0.2cm}
  \label{WIKIPerson}
\end{figure}

\subsubsection{Data Annotation}
The primary goal of WIKIPerson is to link the PERSON mention in the image to the correct Wikipedia entity. As a consequence, the annotators need to identify the person mention and label each mention with the corresponding Wikipedia entity in the form of a Wikidata id.~\footnote{https://en.wikipedia.org/wiki/Wikidata\#Concept} 

In the earlier step, Spacy is used to identify the caption of origin image-text pairs to extract possible PERSON entities. MTCNN is adopted to recognize the faces, supplying bounding boxes in the picture. So the annotators only need to check the faces in the bounding box and choose the corresponding entity from the results generated by searching the keywords of PERON entities detected in the caption. In this way, we can largely reduce the labor in labeling the entity of each mention. Mentions that do not have corresponding entities in Wikipedia will be filtered in the procedure.

In the process of data annotation, we designed end-to-end labeling web demos to facilitate manual annotation. The provided information on the website includes news images, captions, news content, and possible candidate entities with pictures and descriptions to help the annotator make judgments. All annotators have linguistic knowledge and are instructed with detailed annotation principles. The annotators need to link the mention with each bounding box to the correct entity in Wikipedia. Finally, after the labeling, we can get the dataset full of the image which comprises several mentions with each bounding box and corresponding entity WikiId.~\footnote{More examples from WIKIPerson are shown in Appendix.}

\begin{table}[t!]\footnotesize
\setlength\tabcolsep{6pt}%
  \begin{tabular}{l|c|c|c|c}
    \toprule
    & $\#Image$ & $\#E_{cov}$ & $\#M^{I}_{avg}$ & $\#KB$\\
    \midrule
    \midrule
    WIKIPerson & 48K & 13k & 1.08 & 120K \\
  \bottomrule
\end{tabular}
\caption{Statistics of WIKIPerson. $\#E_{cov}$ and $\#M^{I}_{avg}$ denotes number of covered entities and average number of mentions per image, respectively.}
\label{statistic}
\end{table}

\begin{figure}[t!]
  \centering
  \includegraphics[width=0.5\linewidth]{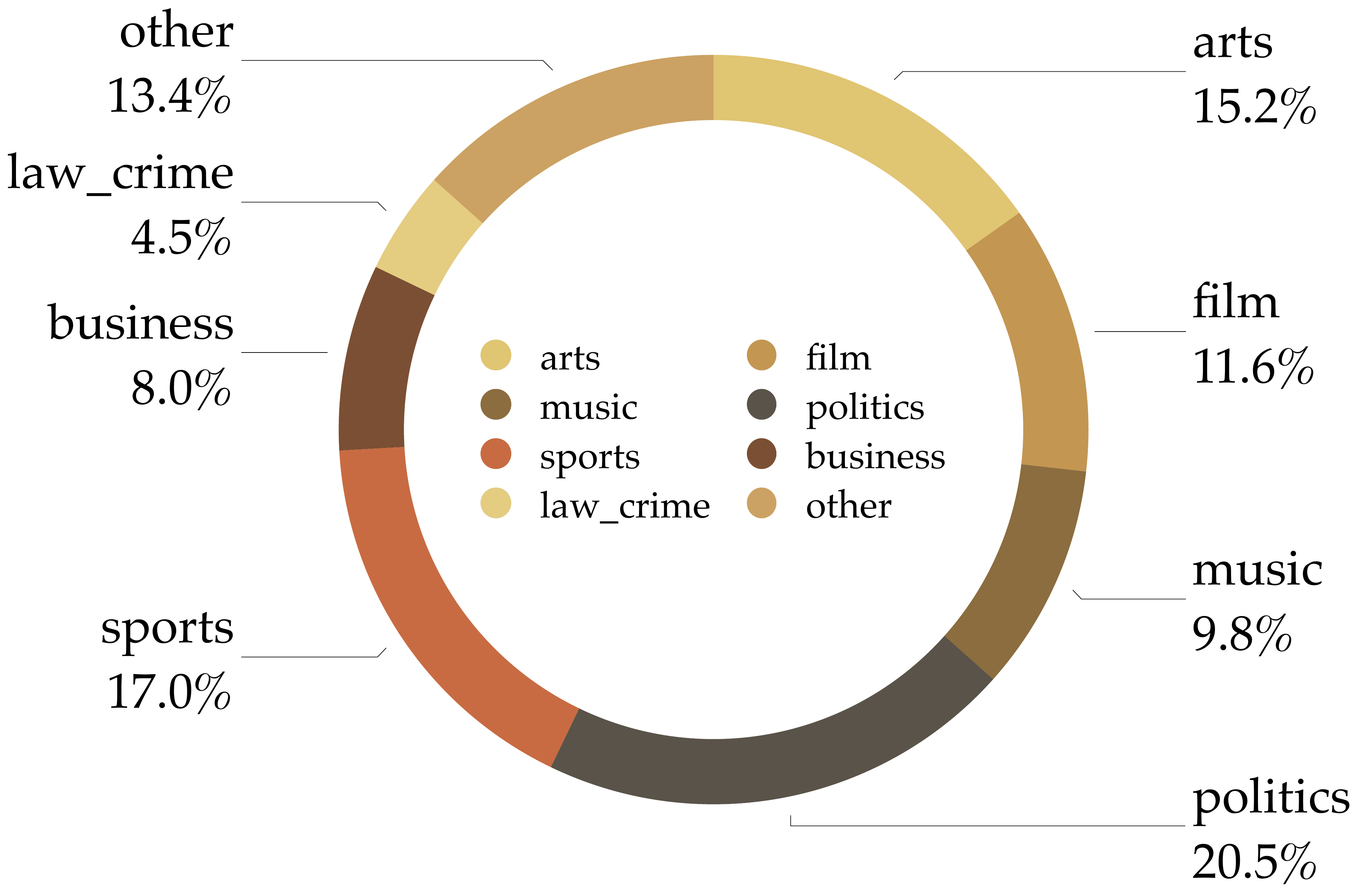}
  \includegraphics[width=0.48\linewidth]{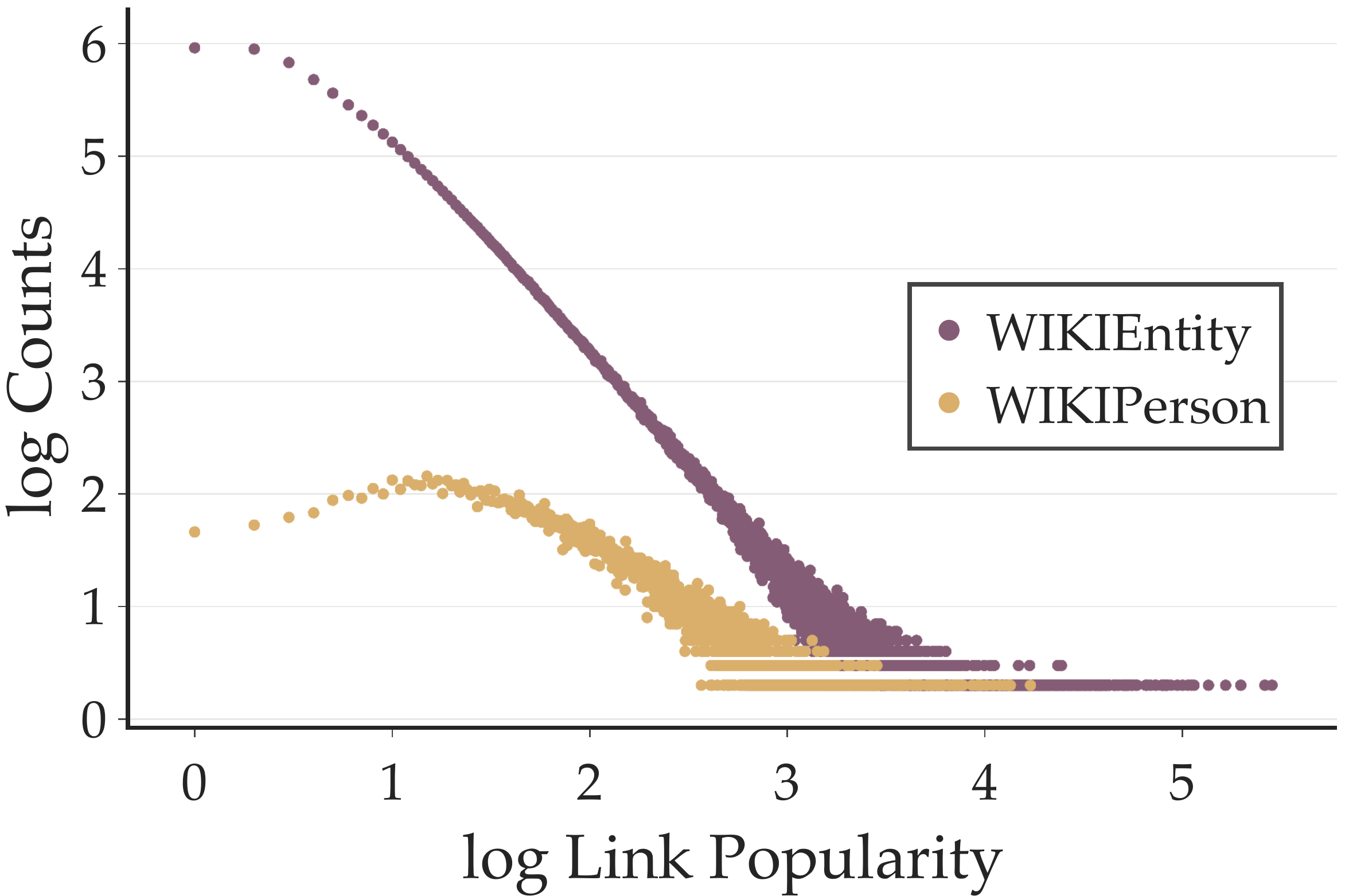}
  \caption{\textbf{Left}: Topic distribution of entities in WIKIPerson. \textbf{Right}: Link popularity distribution between entities in WIKIPerson and the whole Wikipedia.}
  \label{topic}
  \vspace{-0.2cm}
\end{figure}

\subsection{Dataset Analysis}
\subsubsection{Basic Statistics}
Table~\ref{statistic} shows the statistics of the WIKIPerson in detail. The dataset contains a total of 48k different news images, covering 13k out of 120K (i.e.$|E|\approx120K$) PERSON named entities, each of which corresponds to a celebrity in Wikipedia. Many entities appear many times in the data, which ensures that entities can be fully learned. Unlike many datasets in traditional EL, the image of the PERSON named entity usually focuses on a single person in the news except for the scene such as group photo, debate, etc. As a result, the average amount of the mention per image is about 1.08 and only about 3k images contain more than one mention. 

\subsubsection{Entity Distribution}
The WIKIPerson comprises diverse PERSON named entity types such as politicians, singers, actresses, sports players, and so on, from different news agencies. These entities do not belong to a single analogy but are widely distributed in different topics, occupations, skin colors, and multiple age stages. The detailed information is shown in left of Figure~\ref{topic}. It can be observed that in addition to the common politician in the news, the dataset also includes artistic, sports, entertainment, and even criminal topics, which greatly increases the richness of image information. The diversity makes the task could pay attention to the alignment between the background information of the picture, e.g., visual context and entity's meta info in KBs.

Moreover, considering the difference in entities' popularity, we analyzed the link-popularity of the entities in the WIKIPerson compared to that in the whole Wikipedia. As shown in the right of Figure~\ref{topic}, both covered entities and the whole Wikipedia entities conform to the long-tailed distribution, which ensures that the dataset will not be biased because of some significantly popular entities. Generally speaking, celebrities are likely to be reported in news articles, which causes the entity in our dataset to be more prevalent than in the whole Wikipedia. 
To the best of our knowledge, WIKIPerson is the first diverse human-annotated PERSON-entity-aware dataset with high research value. 

\begin{figure}[t!]
  \centering
  \includegraphics[width=\linewidth]{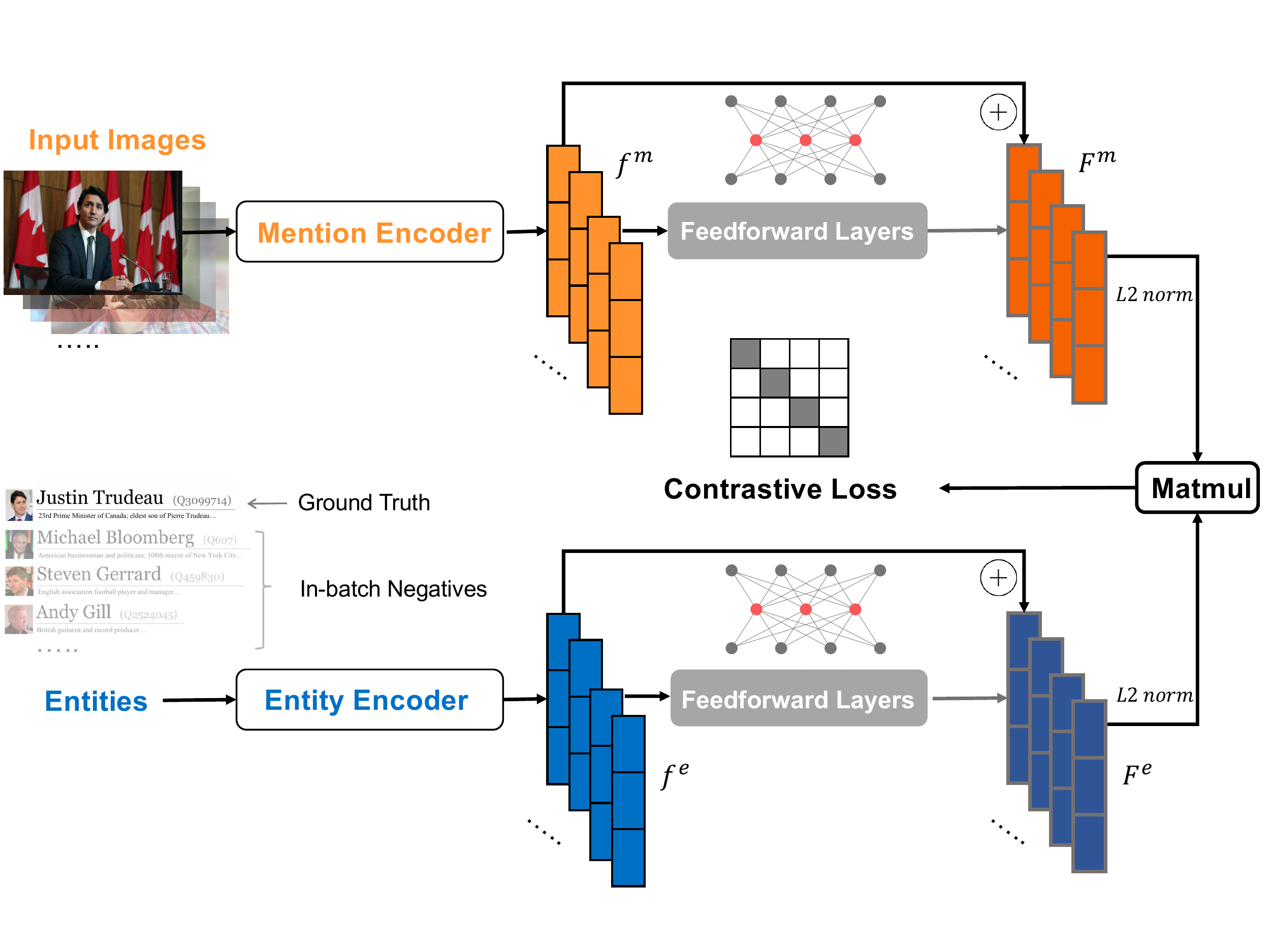}
  \caption{The overall framework of different baselines.}
  \label{model}
  \vspace{-0.5cm}
\end{figure}

\begin{table*}[t!]\small
\setlength\tabcolsep{6 pt}%
\begin{tabular}{ccl|llll|lll}
\toprule 
\hline
\text { } & \text { Sub-Task } & \text { Model } & \multicolumn{4}{c|}{Recall} & \multicolumn{3}{c}{MRR}\\
\text { } & \text {} & \text {} & \text{R@1} & \text{R@3} & \text{R@5} & \text{R@10} & \text{MRR@3} & \text{MRR@5} & \text{MRR@10}\\

\midrule \multirow{7}*{\rotatebox{90}{zero-shot}} & \text {V2VEL} & ResNet & 0.3097  &  0.4053 &  0.4479  & 0.5076 & 0.3518 &  0.3616 & 0.3695\\
~ & \multirow{2}*{V2TEL} & \text{CLIP\_N} & 0.4393 & 0.5673 & 0.6145 & 0.6724 &  0.4964 & 0.5071 & 0.5149\\
~ & ~ & \text{CLIP\_N\_D} & 0.4586 & 0.5872 & 0.6323 & 0.6827 & 0.5158 & 0.5260 & 0.5328\\
~ & \multirow{4}*{V2VTEL} & \text{ResNet+CLIP\_N} & 0.5644 &0.6665 &0.6981 &0.7309 & 0.6101 & 0.6174 & 0.6217 \\
~ & ~ & \text{ResNet+CLIP\_N\_D} & 0.5892& \textbf{0.6859}& \textbf{0.7102} & \textbf{0.7440} & \textbf{0.6327} & \textbf{0.6383} & \textbf{0.6429}
 \\
~ & ~ & \text{CLIP\_N + ResNet} & 0.5667 &0.6618&0.6893&0.7072 & 0.6095 & 0.6158 & 0.6184 \\
~ & ~ & \text{CLIP\_N\_D + ResNet} & \textbf{0.5895} & 0.6794 & 0.7066 & 0.7235 & 0.6302 & 0.6365 & 0.6389 \\
\midrule \multirow{7}*{\rotatebox{90}{fine-tune}} & \text {V2VEL} & ResNet & 0.4212 & 0.5530 & 0.5832 &  0.6428 & 0.4701 & 0.4821 & 0.4899\\ 
~ & \multirow{2}*{V2TEL} & CLIP\_N &  0.5527 & 0.6860 & 0.7250 & 0.7756 & 0.6126 & 0.6215 & 0.6285 \\ 
~ & ~ & \text{CLIP\_N\_D} & 0.5946 & 0.7180 & 0.7550 & 0.8022 & 0.6550 & 0.6634 & 0.6697\\ 
~ & \multirow{4}*{V2VTEL} & ResNet+CLIP\_N &  0.7171 & 0.8115 & 0.8385 & 0.8634 & 0.7600 & 0.7661 & 0.7696 \\ 
~ & ~ & ResNet+CLIP\_N\_D & 0.7301 & \textbf{0.8242} & \textbf{0.8512} & \textbf{0.8798} & 0.7714 & 0.7776 & \textbf{0.7815} \\ 
~ & ~ & CLIP\_N + ResNet & 0.7180 & 0.7921 & 0.8082 & 0.8177  & 0.7502 & 0.7539 & 0.7552\\ 
~ & ~ & CLIP\_N\_D + ResNet & \textbf{0.7370} & 0.8178 & 0.8347 & 0.8445 & \textbf{0.7739} & \textbf{0.7778} & 0.7792 \\ 
 \bottomrule
\end{tabular}

\caption{Experimental results of baselines among three sub-tasks under both zero-shot and fine-tuned settings.}
\vspace{-0.08cm}
  
\label{result}
\end{table*}

\section{Baseline Methods}
Generally, the VNEL task is to link mentions in the input image with the corresponding entities from a large-scale KB. Typically, the existing VNEL system is often implemented as a two-stage process, i.e., the candidate retrieval stage and the entity disambiguation stage, to balance the efficiency and the effectiveness. In this work, we implement a fast end-to-end linking directly from a large-scale collection by employing an efficient model.

We take a widely-used bi-encoder contrastive learning framework to learn robust and effective representations of both visual mentions and entities. 
Given a visual mention $V^m$ and a candidate entity $e_i$, which is accompanied by visual description $V_{e_i}$ and/or textual description $T_{e_i}$, the framework aims to produce a relevance score between the mention and the entity. The overall structure of the framework is shown in Figure~\ref{model}, which consists of two major components, namely the mention encoder and the entity encoder. These two encoders aim to extract features as embeddings $f^m$ for the input image and $f^e$ for the entity. For each encoder, we directly take existing pre-trained models as the implementation. Inspired by existing works~\cite{Clip-adapter, tip-adapter} in applying pre-trained model, we add a feed-forward layer to transform the vector generated from the encoder to the task-oriented embedding space. After that, a residual connection \citep{kaiming} is added to obtain $F^m$ and $F^{e_i}$, followed by using L2 norm and dot-production to calculate the similarity score.

$$
\begin{gathered}
f^m =Encoder^m \left( V^m \right), f^{e_i} = Encoder^{e}\left( e_i  \right)\\
{F}^{m} = f^m + \operatorname{ReLU}\left( f^m \mathbf{W}_{1}^{m}\right) \mathbf{W}_{2}^{m}\\
{F}^{e_i} = f^{e_i} + \operatorname{ReLU}\left( f^{e_i} \mathbf{W}_{1}^{e}\right)\mathbf{W}_{2}^{e}\\
{e^{*}}(m)=\underset{e_{i} \in E}{\arg\max}  {F}^{m}\cdot {F}^{e_i}\\
\end{gathered}
$$

Where $\mathbf{W}_{1}^{m}$ and $\mathbf{W}_{2}^{m}$ are learnable parameters for mention representation learning, and $\mathbf{W}_{1}^{e}$ and $\mathbf{W}_{2}^{e}$ are learnable parameters for entity representation learning.

Since each sub-task of VNEL have different types of inputs, we thus implement each baseline with different encoders:
\begin{itemize}
    \item \textbf{V2VEL Encoders}: We adopt ResNet \citep{ResNet} in a single-modal way following \citep{facenet}, which has been pre-trained on the vggface2 \citep{VGGface} to extract visual features. Here, both mention and entity encoder use ResNet and share the parameters.
    \item \textbf{V2TEL Encoders}: We directly take CLIP \citep{CLIP}, which has been pre-trained with a large-scale image-text dataset, to implement the mention encoder and the entity encoder. For entity encoder, we apply two types of textual information about entity, i.e., entity name (CLIP\_N) and entity name with description (CLIP\_N\_D), to study the influence of the entity's meta info.
    \item \textbf{V2VTEL Encoders}: We combine encoders of V2VEL and V2TEL to implement the V2VTEL. Specifically, we take a simple but effective strategy that uses one model to recall Top-K results and the other to re-rank. For example, ResNet + CLIP means recall with the ResNet first and re-rank Top-K results with CLIP again. We also test different combinations about the order of V2VEL encoders and V2TEL encoders, whose results are listed in Section 4.1. \footnote{The detailed analysis of this strategy is displayed in the Appendix due to the page limitation.}
\end{itemize}
In the training step, the contrastive loss function of a single mention-entity sample is defined as:

$$
\begin{small}
\begin{gathered}
\mathcal{L}\left(V^m, e_i \right)=-\log \left[\frac{\exp \left( \Phi\left(V^m, e_i^{+}\right)/ \tau\right)}  { \sum^{-}   +\exp \left(  \Phi\left(V^m, e_i^{+}\right)/ \tau\right)}\right]\\
\sum^{-}= {\sum_{k \neq i} \exp \left(\Phi\left(V^m, e_k^{-}\right) / \tau\right)}    
\end{gathered}
\end{small}
$$

where $e_i^{+}$ represents the ground truth positive entity of $V_m$ and $e_k^{-}$ denotes the $k^{th}$ candidate of $V^m$ in the batch, which is all negative samples. $\tau$ is the temperature coefficient that helps control the softmax's smoothness\cite{temp}. 

\section{Experiments}
During experiments, we split images in WIKIPerson into train, dev, and test set with the ratio of 6:2:2. Besides, to avoid the bias of popular entities affecting the evaluation, each named entity appears at most once in test set. For evaluation, we report two widely-used metrics of Top-k retrieve: Recall@K (K=1, 3, 5, 10) and Mean Reciprocal Rank (MRR@K, K=3, 5, 10).\footnote{The description about the evaluation metrics can be found in the Appendix.}

\subsection{Results}
All results are summarized in Table~\ref{result}. Since all the encoders we adopt are pre-trained and can be directly applied in each task, we thus report both zero-shot and fine-tuned performances to show the effectiveness of all baselines.

\textbf{Zero-shot v.s.~fine-tune.} In zero-shot, we directly use the embedding generated from the encoder as the feature. As we can see, ResNet has achieved a reasonable good performance for R@10 (i.e, 0.5076), which demonstrates the effectiveness of the pre-trained model. Moreover, we can see that the CLIP, which is pre-trained with about 400M image-caption pairs, has achieved better performances against ResNet with either CLIP\_N or CLIP\_N\_D across all metrics.
When combining ResNet with CLIP, we observe a distinct improvement for all combination, which demonstrate the effectiveness in combining both visual description and textual description in VNEL.
While comparing the zero-shot with fine-tuned baselines, all models have obtained significant improvements, e.g., an average improvement of MRR@10 is 0.13. The improvements verify the quality of dataset and demonstrates that the WIKIPerson could significantly boost the ability of visual named entity linking.

\begin{figure*}[t!]
  \centering
  \includegraphics[width=\linewidth]{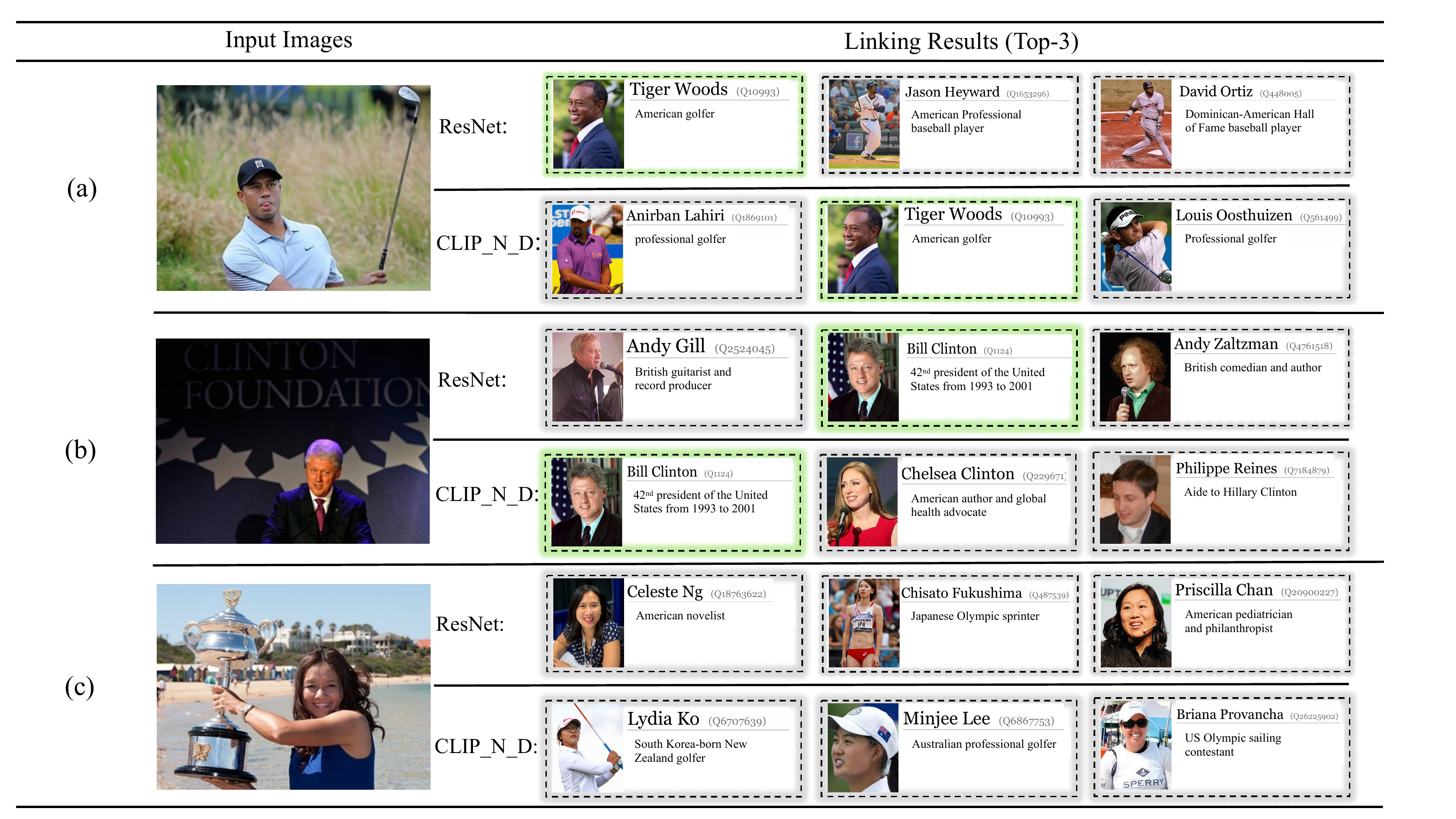}
  \caption{The qualitative case studies of Top-3 predicted entities. The result with a green border is the ground truth entity of the input image.}
  \label{case}
\end{figure*}

\textbf{Sub-tasks of VNEL.}
We focus on the below part of table \ref{result} where all models are fine-tuned on WIKIPerson. 

1) The V2VEL sub-task: As the most fundamental part concerning VNEL, the ResNet extracts features for both visual mentions and visual descriptions of entities, and matches them in visual feature space. However, it obtains generally low absolute numbers in different evaluation metrics, e.g., $0.4212$ on R@1, which leaves a large room for improvement. A possible reason is that the image of an entity in KB are often earlier pictures which show very different state (e.g., age and occasion) with entities appeared in news articles. 

2) The V2TEL sub-task: CLIP obtains higher performance compared to ResNet by matching the visual mention with textual descriptions of the entity. Besides, these results show that the cross-modal matching between the image and the text is very powerful in linking images with entities. Moreover, by comparing the two different types of textual information about the entity, we can see that entity description could provide useful information in distinguishing disambiguate entities since CLIP\_N\_D outperforms CLIP\_N over all metrics.

3) The V2VTEL sub-task: By combining the textual information and visual information of each entity, the performance could be further boosted. For example, the relative improvement of ResNet+CLIP\_N\_D over ResNet and CLIP\_N\_D against R@1 is about $73\%$ and $23\%$, respectively. These results verify that both textual and visual modality of the entity could complement each other in linking visual mentions with named entities. Moreover, as for different order of combination between ResNet and CLIP, we can see that each method could obtain a relatively close performance, which confirms the effectiveness of the strategy in combining the V2VEL method and the V2TEL method. 

\subsection{Qualitative Analysis}
To better understand baseline methods among different sub-tasks, we show several cases in Figure~\ref{case}. The input image is on the left, and the top 3 predicted results are partitioned into two rows corresponding to different baselines. The entity with a green border is the ground truth entity.

The first case of Figure~\ref{case} is a picture of a famous American golfer named Tiger Woods. ResNet could identify the correct entity, and other returned results have a similar face to the input image. CLIP\_N\_D also returned the ground truth entity at the second position in the top-3 results, and all three candidates are professional golfers. This shows that only text descriptions may unable to disambiguate between the correct entity and irrelevant entities.

Analogously, the second case is an image about Bill Clinton speaking at his foundation. ResNet links it to the entity "Andy Gill", which looks very similar to Clinton. While CLIP\_N\_D correctly predicts the ground truth entity in the first position, and all returned entities are related to Clinton. This verifies that CLIP\_N\_D can learn high-level association between image mention and entity meta-info.

The last case is an image of a famous Chinese tennis sports player named Li Na. We can see that the image has complicated backgrounds, and both ResNet and CLIP\_N\_D cannot link the mention with the ground truth entity in top-3 returned results. This motivates the need for focused research on building effective VNEL models.

From all the above cases, it is clearly presented that ResNet pays more attention to the pixel-level matching, and CLIP learns high-level semantic connection between mentions and entities. However, the dynamic nature of the input images highlights the difficulty of the task, especially for entities with outdated pictures. We believe this work could pave the way for better visual entity linking.

\section{Related Work}

\textbf{Entity Linking}. There is extensive research on EL, which serves as a classic NLP task. With the help of large-scale pre-train language models \citep{Bert,Roberta}, several recent deep learning methods \citep{EL_g_1,EL_g_2,EL_g_3} achieve 90\%+ accuracy on AIDA \citep{AIDA}, which is a commonly used high-quality robust EL dataset. However, as mentioned in \citep{mention_early}, it seems that the current methods have already torched the task ceiling. As a result, many more challenging EL-related tasks are formulated. For example, zero-shot entity linking \citep{Zero-shot_EL_1,Zero-shot_EL_2}, engaging other features like global coherence across all entities in a document, NIL prediction, joining MD and ED steps together, or providing completely end-to-end solutions to address emerging entities is rapidly evolving \citep{EL_survey}.

\textbf{Multi-modal Entity Linking}. Recently, Multi-modal Entity Linking(MEL) \citep{snap} task has also been proposed for consideration. Given a text with images attached, MEL uses both textual and visual information to map an ambiguous mention in the text to an entity in the KBs. \citep{snap} proves that image information helps identify the mention in social media for the fuzzy and short text. Furthermore, \citep{Twitter} transfer the scene to Twitter and perform MEL on Twitter users. \citep{Weibo} proposes an attention-based structure to eliminate distracting information from irrelevant images and builds a multi-source Social Media multi-modal dataset. \citep{WikiDiverse} builds a multi-modal Entity Linking Dataset with Diversified Contextual Topics and Entity Types. However, for all those works, the text input plays a vital part, and the visual input only serves as a complementary role to the text.

\textbf{Multi-modal Dataset}. At the same time, our work is also related to the multi-modal image-text datasets, which is also a hot issue in recent years. Flicker30k \citep{Fliker} annotates 30k image-caption pairs from Flicker with five descriptive sentences per image, such as "a man is wearing a tie." In addition, MSCOCO caption \citep{coco} scale up the size with over one and a half million captions describing over 330000 images. However, the caption in all these datasets is descriptive sentences and non-entity aware. As a result, some work has started to build a news-related dataset for entity-aware image caption tasks. For example, \citep{BreakingNews} focus on the news website and have crawled 100k image-caption pairs. \citep{GoodNews,VisualNews} expand the size of the dataset. Nevertheless, the detailed entity information is neither annotated nor linked to the KBs. 

\section{Conclusion and Future Work}
To tackle the limitation that previous visual entity linking either rely on textual data to complement a multi-modal linking or only link objects with general entities, we introduce a purely Visual-based Named Entity Linking task, where the input only contains the image. The goal of this task is to identify objects of interest in images and link them to corresponding named entities in KBs. Considering the rich multi-modal contexts of each entity in KBs, we propose three different sub-tasks, i.e. the V2VEL sub-task, the V2TEL sub-task, and the V2VTEL sub-task. Moreover, we build a high-quality human-annotated visual person linking dataset, named WIKIPerson, which aims at recognizing persons in images and linking them to Wikipedia. Based on WIKIPerson, we introduce several baseline algorithms for each sub-task. According to the experimental results, the WIKIPerson is a challenging dataset worth further explorations. In the future, we intend to build a larger scale VNEL dataset with diverse types and adopt more advanced models to achieve higher accuracy.

\section*{Limitations}
\textbf{Low extensibility of the entity information.} In the V2VEL sub-task, each entity in the KB can have more than one attached image. However, in our paper, only the first image is selected for convenience, which will inevitably omit additional information. At the same time, in the V2TEL sub-task, we only use the short descriptive sentences of the entity. How to integrate longer unstructured text information is also a problem worth exploring.

\section*{Ethics Statement}
We collected data based on open-source datasets and databases. These data have been strictly manually reviewed and do not contain any pictures that are sexual or violate politics.
We are authorized by the relevant authority in our university to hire employees from the laboratory to build the platform and carry out the annotations. All employees are adults and ethical. On average, they were paid £5–£10/hour.

\section*{Acknowledgements}
This work was funded by the National Natural Science Foundation of China (NSFC) under Grants No. 61902381 and 62006218, the Youth Innovation Promotion Association CAS under Grants No. 2021100 and 20144310, the Young Elite Scientist Sponsorship Program by CAST under Grants No. YESS20200121, and the Lenovo-CAS Joint Lab Youth Scientist Project.

\bibliography{anthology,custom}
\bibliographystyle{acl_natbib}

\clearpage

\appendix

\section{Baselines Details}

\textbf{Parameters Setting.} In the architecture, we set the number of layers in the feed-forward as 2 and the dimensions are [512*1024, 1024*512] both for mention and entity in the two models. The initial learning rate is set to 2e-4 for ResNet and 2e-6 for CLIP. Images are all resized to 224 × 224 pixels according to the common size and textual information is truncated to 77 words. The batch sizes for ResNet and clip are both set to 64. All the methods are implemented in Pytorch \citep{torch} and optimized by the AdamW \citep{adamw} algorithm.

\textbf{Experimental setup.}
We train our models on two NVIDIA Tesla V100 GPU. We train each model with much to 20 epochs. For inference, we use Faiss\footnote{https://github.com/facebookresearch/faiss} to achieve fast recall in large-scale embedding space with about 500ms per instance.

\section{Evaluation Metrics}
All evaluation and empirical analysis are reported by two widely-used metrics of Top-k retrieve: Recall and Mean Reciprocal Rank (MRR). The final result is the average score among all the cases.

$$
\begin{gathered}
\mathrm{Recall@K}=\frac{1}{Q} \sum_{i=1}^{Q} \mathbf{1}_{qk_i}(gt_i)\\
\mathrm{MRR@K}=\frac{1}{Q} \sum_{i=1}^{Q} \frac{1}{\operatorname{rank}_{i}}
\end{gathered}
$$

where $\mathbf{1}_{A}(x)$ denotes a {0,1} valued indicator function. ${qk_i}$, $gt_i$ are the Top-k result and the ground truth of query i. MRR is a measure to evaluate systems that "Where is the first relevant item". For a single query, the reciprocal rank is $\frac{1}{\text { rank }}$ where rank is the position of the highest-ranked answer. If no correct answer was returned in the query, then the reciprocal rank is 0.

\begin{figure}[t!]
  \centering
  \includegraphics[width=\linewidth]{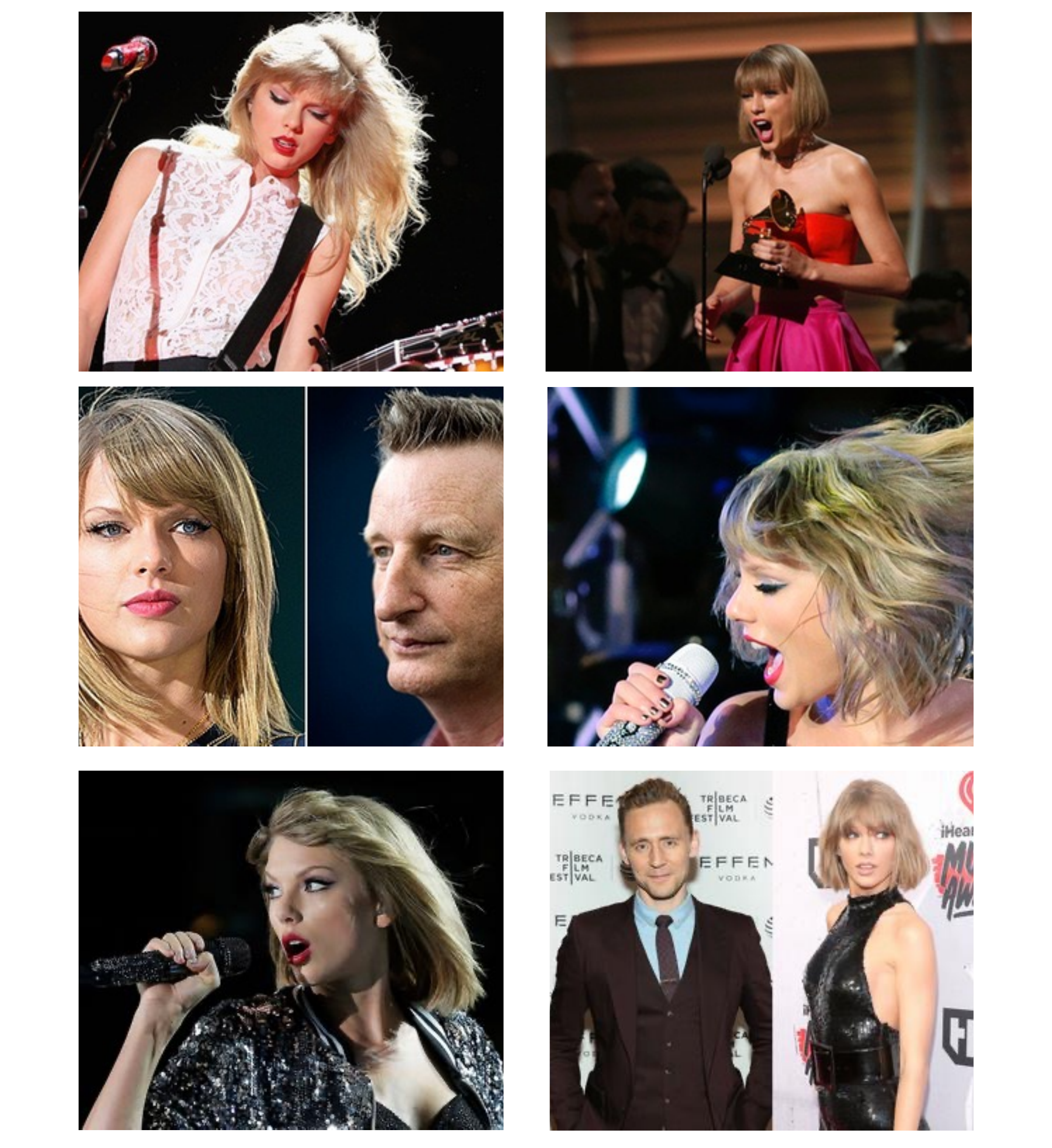}
  \caption{The images of Taylor Swift (Q26876, a famous American singer-songwriter) in WIKIPerson.}
  \label{f_more_1}
\end{figure}

\begin{figure}[t!]
  \centering
  \includegraphics[width=\linewidth]{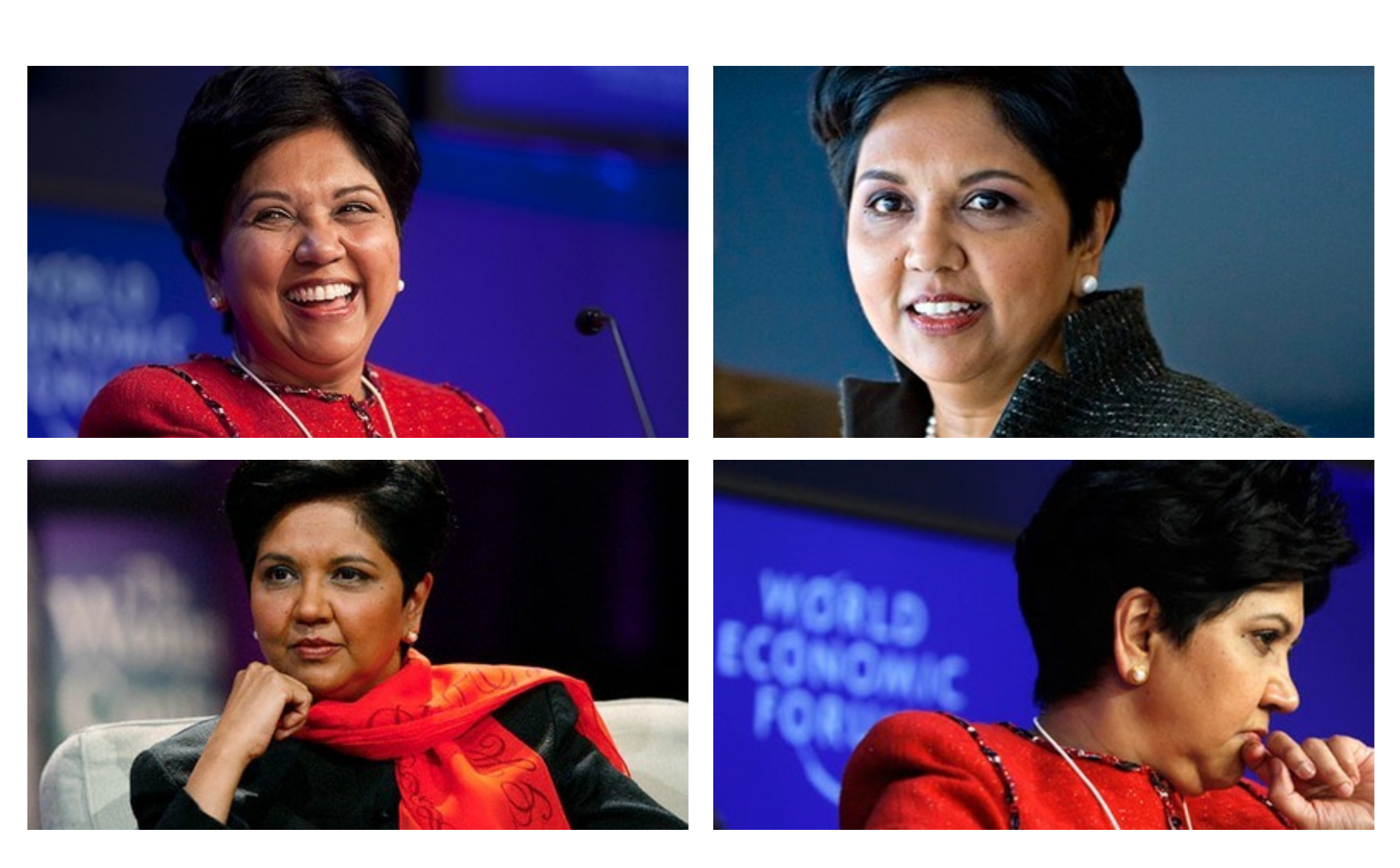}
  \caption{The images of Indra Nooyi (Q264913, Indian American business executive and former CEO of PepsiCo) in WIKIPerson.}
  \label{f_more_2}
\end{figure}

\section{More Examples from WIKIPerson}
To demonstrate more details of our dataset, we pick two examples from our dataset. (Figure~\ref{f_more_1}, Figure~\ref{f_more_2}).

\section{Detailed Analysis}

According to the experimental results, the re-ranking strategy improves performance to a certain degree. So we conduct a detailed analysis of the strategy to help understand the reason and provide some insights for future model designs.

\begin{figure}[t!]
  \centering
  \includegraphics[width=0.95\linewidth]{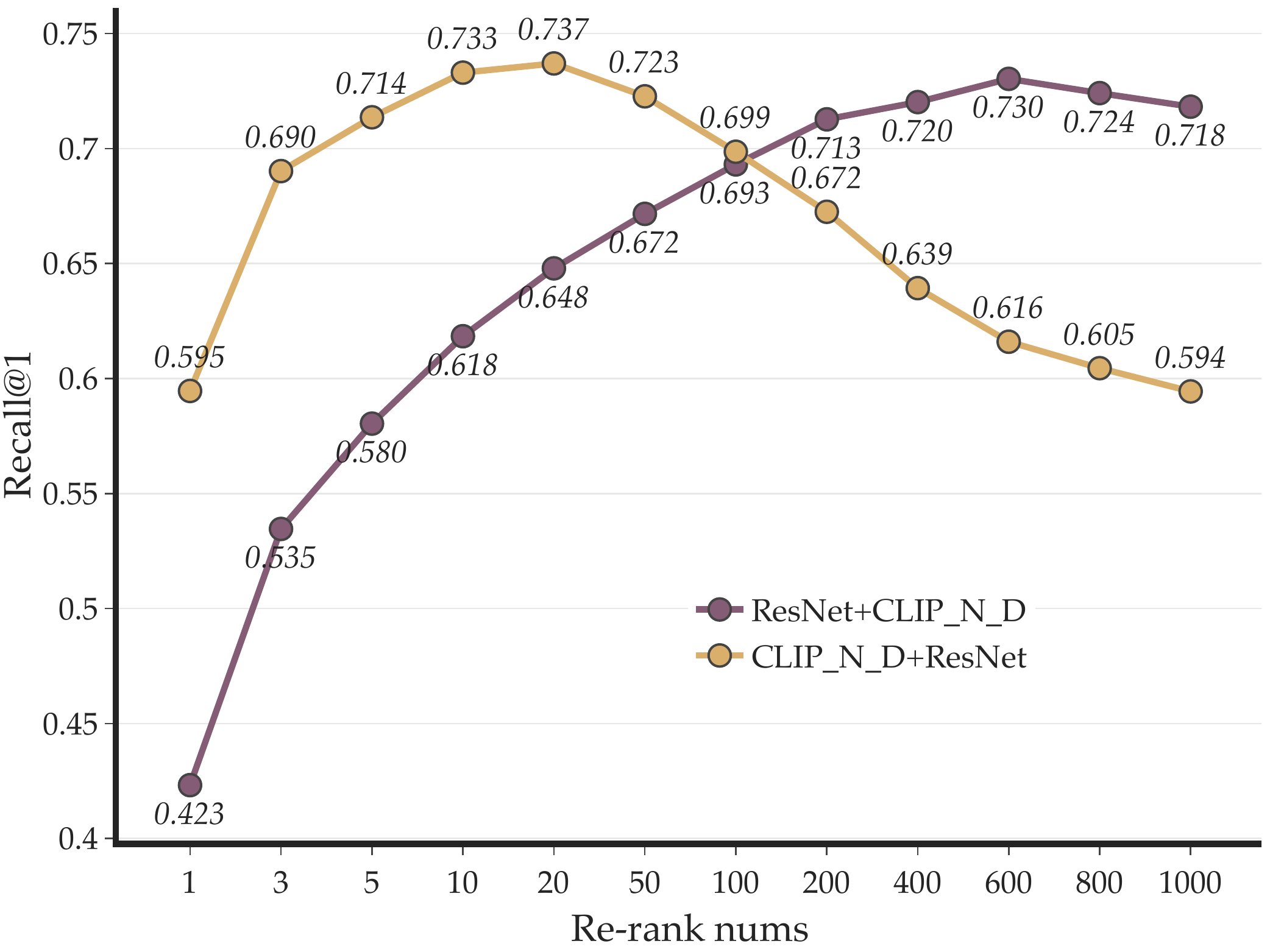}
  \caption{The Recall@1 of the models with different re-rank size.}
  \label{re-rank}
\end{figure}

Firstly, we analyze the effect of re-ranking sequence length, which is the main factor affecting the result. Specifically, we conduct research on the re-ranking sequence length. Then we plot the Recall@1 for ResNet + CLIP\_N\_D and CLIP\_N\_D + ResNet in Figure~\ref{re-rank}. From the results, we can see that both two methods achieve high performance as the re-ranking length increases at the beginning. Then it starts to decrease slightly. It can be simply inferred that when the re-ranking length continues to grow to the size of the |E|, the re-ranking model can be equal to the single Reset or CLIP\_N\_D. Besides, these two models have different Inflection Points and speeds of the downtrend. CLIP\_N\_D + ResNet reaches its peak at lower re-rank length and decent sharply while ResNet + CLIP\_N\_D increases until re-rank length equals 600 and decent slowly. The reason for the phenomenon is that  CLIP\_N\_D outperforms ResNet. As a result, a larger re-rank size is necessary for ResNet to guarantee to recall the ground truth.

\begin{figure}[t!]
  \centering
  \includegraphics[width=0.95\linewidth]{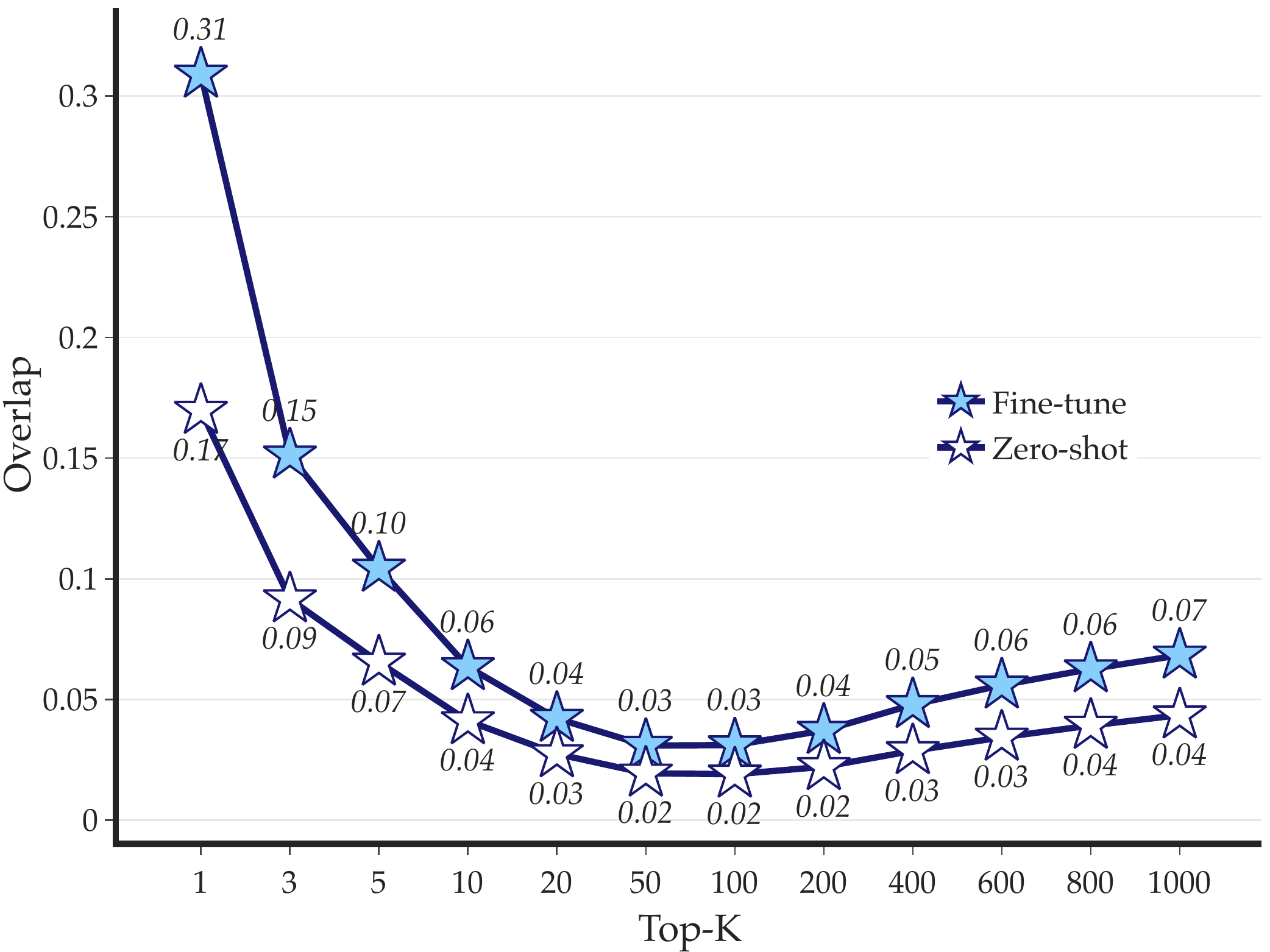}
  \caption{Overlap of Top-k result between CLIP\_N\_D and ResNet in zero-shot and fine-tune.}
  \label{overlap}
\end{figure}

Secondly, we notice that the Top-k results of CLIP\_N\_D and ResNet differ greatly. As a result, we plot the precise overlap between ResNet and CLIP\_N\_D's Top-k result in Figure \ref{overlap}. 

The origin and fine-tune model have the same trend: with the increase of the K, the overlap decreases first and increases later. When k nears 50, the overlap minimum. For fine-tune model, it has a higher overlap than the zero-shot. The overlap starts from 30.1\%, which means only the 30.1\% of entities are identical among the Top-1 results between the two models even though they have comparable performance. Then it drops to  15\% sharply. When k equals |E|, the overlap will reach 100\%. Smaller coverage with high and comparable model performance ensures that using one model to re-ranking based on the recall of the other model could improve performance significantly.
\end{document}